\documentclass[acmsmall,nonacm,screen]{acmart}

\usepackage{booktabs}
\usepackage{xurl}
\usepackage{listings}
\usepackage{placeins}
\setcopyright{none}
\renewcommand\footnotetextcopyrightpermission[1]{}

\makeatletter
\newcommand{\nocountry}{\@ACM@countrypresenttrue\@ACM@citypresenttrue}
\makeatother

\begin{document}

\title{Budget, Readout or Reasoning? Where Small Reasoning Models Lose Value in Auction Bidding}

\author{Mahbod Khaleti}
\email{mahbod.kh2005@gmail.com}
\affiliation{\institution{Dataism Laboratory for Quantitative Finance}\nocountry}

\author{Seyed Reza Tavakoli}
\email{seyedreza.shiyade@gmail.com}
\affiliation{\institution{Dataism Laboratory for Quantitative Finance}\nocountry}

\author{Baran Hoseini}
\email{baranhoseini.cloudstear@gmail.com}
\affiliation{\institution{Independent researcher}\nocountry}

\author{Alireza Noori}
\affiliation{\institution{Dataism Laboratory for Quantitative Finance}\nocountry}

\author{Amir Amini}
\affiliation{\institution{Princeton University}\nocountry}
\affiliation{\institution{Dataism Laboratory for Quantitative Finance}\nocountry}

\author{Koorosh Aslansefat}
\email{k.aslansefat@hull.ac.uk}
\affiliation{\institution{University of Hull}\city{Hull}\nocountry}

\author{Marjan Alirezaie}
\email{marjan.alirezaie@torontomu.ca}
\affiliation{\institution{Toronto Metropolitan University}\city{Toronto}\nocountry}

\author{Ali Habibnia}
\authornote{Corresponding author: habibnia@vt.edu}
\email{habibnia@vt.edu}
\affiliation{\institution{Virginia Tech}\nocountry}
\affiliation{\institution{Dataism Laboratory for Quantitative Finance}\nocountry}

\thanks{Code, data and experiment outputs are available at \url{https://github.com/DataismLab/bid-reasoning}.}

\authorsaddresses{}
\renewcommand{\shortauthors}{Khaleti et al.}

\begin{abstract}
Ad platforms and marketplaces show bidders a bid landscape, the estimated chance of winning at each candidate bid, and increasingly let agents built on reasoning language models choose the bid. We give six open reasoning models of 4B to 14B parameters such landscapes, with win curves from eBay closing prices and from a simulated advertising market (AuctionNet), and score their bids on held-out auctions against an exact calculator. Because the agent is handed the estimates, the lost value can be split by cause. Two failures belong to the models. They pass over the best option even when the values written in their own trace already rank it first, and their wrong answers lean toward the highest bid on the menu. The lean is about the bid's size, not its place in the list, and it survives sampling. At 768 forced tokens on eBay, under pay-your-bid pricing, the wrong answers cost bidders 7--17\% of the surplus the calculator's bids earn. Arithmetic errors are rare. Two further losses come from operator choices. A token cap cuts traces short: allowed to stop on their own within 4{,}096 tokens, five models recover 91--100\% of the calculator's lead over a model that does no reasoning. Asking for the answer while the reasoning block is still open misreads it: closing the block first lifts the weakest model from 42\% to 84\% on eBay and from 24\% to 81\% on AuctionNet. Finally, an audit that flags bids disagreeing with the trace's own numbers, and overrides them with the calculator's bid, raises the share of the calculator's lead recovered at 768 forced tokens from 42--85\% to 84--100\% on eBay under nested cross-validation, and the same reader settings transfer from eBay to AuctionNet.
\end{abstract}



\maketitle

\section{Introduction}

Much of the web's commerce is priced by automated auction, and platforms increasingly hand the bidder the forecast. Ad platforms and marketplaces expose a bid landscape or win-rate simulator, the estimated chance of winning at each candidate bid, and leave the choice of bid to the advertiser, seller or buyer, or to an agent acting for them. These choices are made per keyword, campaign or high-value listing rather than per impression, so an agent can afford to spend seconds on each. LLM agents are being built to bid, price and negotiate in such markets \cite{chen2024aucarena,zhu2024synthetic,ahmed2026bazaar}, increasingly on reasoning models that spend tokens before acting. That raises two questions for whoever runs the agent: how much reasoning to pay for, and how to tell when the agent's bid can be trusted.

This paper studies both on the simplest version of the decision. The agent sees a private value $v$ and a short menu of candidate bids, each with an estimated win probability $\hat p(b)$, and must pick the bid with the largest expected profit $\hat p(b)(v-b)$. The task is deliberately easy, and that is the point. Because the agent receives the estimates, an exact calculator solves it, so the lost value can be split between the agent's reasoning, its answer, and our forcing and readout protocol. Richer bidding tasks have no such ground truth.

\paragraph{Why not let the calculator bid?}
Deployed agents do not face a calculator-sized problem. The landscape is one input among others the calculator cannot weigh, such as budget pacing, brand goals or the advertiser's notes, and the operator is often not the agent's author: a company running a vendor's bidding agent, or a platform hosting third-party agents, can read the agent's reasoning but cannot replace it. An agent that loses value on the one step a calculator can check is unlikely to do better once that step sits inside a richer decision, so we expect its loss here to be a lower bound, although we do not test richer tasks, and ask which bids an operator should flag for review or override.

\paragraph{What we find}
We run six open reasoning models (Phi-4-mini-reasoning, Qwen3-4B, -8B and -14B, DeepSeek-R1-Distill-Llama-8B and Granite-3.2-8B-Instruct) on 216 bid-landscape scenarios built from eBay closing prices and 216 built from AuctionNet, a simulated advertising market \cite{su2024auctionnet}. We force exact reasoning budgets from 0 to 768 tokens, also let the models stop on their own, and parse every trace with code to find where the gap to the calculator opens.
\begin{itemize}
\item \textbf{Model-side failures.} Most of the loss comes after the best option's value is already on the page. Models pass over the best option even when the values written in their trace rank it first, and their wrong answers lean toward the menu's highest bid. The lean is about size rather than list position (conditional-logit odds ratio 3.9 on both markets for DeepSeek-8B and the Qwen models, 2.2--2.3 across all six), and it survives sampling and a corrected readout. At 768 forced tokens on eBay, wrong answers cost 7--17\% of the calculator's surplus under pay-your-bid pricing, and for DeepSeek-8B and the Qwen models more than half of that loss comes from choosing the highest bid.
\item \textbf{Arithmetic is rare.} For five of six models arithmetic errors are at most 5\% of the gap (17\% under a deliberately pessimistic bound). Granite-8B is the exception.
\item \textbf{Operator-side losses.} A token cap cuts traces short: the Qwen models and DeepSeek-8B lose mainly to the cap and recover 99--100\% of the gap when allowed to stop within 4{,}096 tokens. The readout prompt matters too: asking for the answer while the reasoning block is open misreads Phi-mini's choice, and closing the block first lifts it from 42\% to 84\%.
\item \textbf{A cheap audit.} A text reader that recovers the trace's own per-option numbers, used only to flag disagreement with the answer, raises gap recovered at 768 tokens from 42--85\% to 84--100\% when flagged bids are overridden with the calculator's bid. Its settings chosen on eBay transfer unchanged to AuctionNet.
\end{itemize}

\paragraph{Contributions}
(1) A bid-landscape task with exact ground truth on two markets, scored on held-out auctions, with forced and unforced reasoning budgets. (2) A code-only decomposition of each model's gap into four causes, with lower and upper bounds on arithmetic error. (3) An analysis of which wrong bids models choose, why (size versus position), and what they cost. (4) A readout ablation that separates protocol artifacts from model behavior. (5) An audit that needs no calculator to raise a flag, evaluated out of sample. All code, saved decisions and analysis scripts are released.

\paragraph{Setting}
Our target is a pay-your-bid auction in which an agent sets the bid for a keyword, a campaign or a high-value listing from a landscape; first-price rules have spread in display advertising \cite{despotakis2021firstprice}. eBay itself runs proxy bidding, which we do not model: its closing prices only supply realistic empirical shapes for the win curves. AuctionNet's auctions are generalized second-price, which we also do not model; we use its competing prices in the same way (Section~\ref{sec:auctionnet}).

\section{Related work}

\paragraph{Test-time compute and its failure modes}
Spending more tokens before answering improves accuracy on many tasks \cite{wei2022chain,snell2024scaling}. Our forcing of exact budgets resembles budget forcing \cite{muennighoff2025s1}. Reasoning models can also overthink easy problems \cite{chen2024overthinking} or abandon promising lines too early \cite{wang2025underthinking}. We measure what a budget buys in dollars, and where the remaining value goes.

\paragraph{Faithfulness and answer extraction}
Multiple-choice answers depend on option order \cite{zheng2024mcq,pezeshkpour2024order}, and stated reasoning need not determine the answer \cite{turpin2023unfaithful,lanham2023faithfulness,chen2025reasoningsay}. Our readout ablation is a concrete instance: the same trace yields different answers depending on whether the reasoning block is closed before the answer is requested. Passing over a computed best option is a direct, dollar-valued case of an answer departing from the reasoning.

\paragraph{Tools}
Tool-using models can offload arithmetic \cite{schick2023toolformer}. Our calculator is the ideal tool call; we do not test a model that calls a tool and can fail to.

\paragraph{Bid landscapes and auto-bidding}
Bid landscapes are forecast from auction logs \cite{cui2011bidlandscape} and feed auto-bidding systems \cite{aggarwal2024autobidding}. LLMs now act as real-time bidding agents \cite{cai2025rtbagent} or reasoning modules inside auto-bidding \cite{li2026lbm}, and \citet{duetting2024mechanism} design auctions among LLM agents. AuctionNet \cite{su2024auctionnet} provides a large simulated advertising market for decision-making research.

\paragraph{LLMs as economic agents}
LLM agents have been studied as bidders and economic actors in arenas and benchmarks \cite{chen2024aucarena,zhu2024synthetic,ahmed2026bazaar,raman2024steer,fish2025econevals}. EconEvals \cite{fish2025econevals} and STEER \cite{raman2024steer} score whether decisions are rational. We instead locate where a trace loses value it has already computed, and separate model behavior from protocol choices.

\section{Task and data}

\subsection{eBay scenarios}
We use 628 eBay auctions of Palm Pilot PDAs, Xbox consoles and Cartier watches collected by \citet{jank2010modeling}, in the Kaggle copy \cite{onlineauctionsdataset}, split 70/30 within product into a construction sample and a held-out sample of 189 auctions. Per product, 18 private values $v$ are spaced evenly from the 35th percentile of construction-sample closing prices to 1.25 times the 95th. For each value and menu size $K\in\{5,7\}$ we draw two candidate sets of \$5-step bids below $v$ within the 2nd--98th price percentiles, either as a random contiguous window or as a random sample, never conditioned on the optimum. This gives $3\times18\times2\times2=216$ scenarios. Each bid is shown with $\hat p(b)$, the share of construction-sample closing prices at or below $b$, rounded to four decimals; the chosen bid is scored with the same estimator $p(b)$ on the held-out auctions. These are realistic empirical win curves rather than first-price win rates, and the models and the calculator see the same landscape.

\subsection{AuctionNet scenarios}
\label{sec:auctionnet}
To test a second market we rebuilt the task on AuctionNet \cite{su2024auctionnet}, a simulated advertising market in which 48 auto-bidding agents compete in a three-slot generalized second-price auction over ad opportunities generated by a deep generative model. The data are simulated, not real auction logs. The competing price of an opportunity is the third-highest bid among the 48 agents, the lowest bid that wins a slot. Because this price does not depend on an advertiser's category, we define three products as terciles of the opportunity's conversion probability (high-, mid- and low-intent impressions). Prices, values and bids are shown per 1{,}000 impressions. Win curves are estimated on periods 7--9 (about 500{,}000 opportunities per product) and bids are scored on period 10 (132{,}000--184{,}000 per product). The 216 scenarios are built exactly as on eBay, with a bid step of 2. As on eBay, a bid is scored pay-your-bid, $p(b)(v-b)$, with $p(b)$ the share of held-out opportunities whose competing price is at most $b$; we do not model GSP payments or slot exposure.

\subsection{Synthetic scenarios}
A synthetic version uses the same protocol in the standard uniform first-price setting \cite{krishna2010auction}: 216 scenarios with integer bids against two rivals with uniform values who bid their equilibrium, so payoffs are exact. We use it only to choose and test the audit's settings out of domain.

\subsection{Measures}
The \emph{utility ratio} of a bid $\hat b$ is $q=U(\hat b)/\max_b U(b)$, with expected profit $U(b)=p(b)(v-b)$ under the held-out probabilities and the maximum taken over the menu. The calculator returns $\arg\max_b \hat p(b)(v-b)$ on the displayed estimates; on eBay it scores $q=0.981$ and random choice scores 0.711, and on AuctionNet 0.976 and 0.776.

Our primary measure is the \emph{gap recovered} at budget $c$,
\[
\mathrm{GR}(c)=\frac{q(c)-q(0)}{q_{\text{calc}}-q(0)},
\]
averaged per decision. $q(0)$ is the same model, prompt and readout with no reasoning tokens, so $\mathrm{GR}$ isolates what the tokens buy, and $\mathrm{GR}=1$ means the model matches the calculator. Because $q(0)$ differs across models, $\mathrm{GR}$ does not rank models by the quality of their bids; Tables~\ref{tab:qbudget} and~\ref{tab:anmain} report $q$ itself. Pooling dollars over decisions instead weights the most expensive product most (Cartier on eBay) and is reported as secondary. Intervals are 95\% percentile intervals from 2{,}000 bootstrap draws that resample the 18 value families within each product (on AuctionNet, within each category); a product-level bootstrap is reported in Section~\ref{sec:robust}.

\paragraph{A second scoring curve}
For robustness we also score every bid against rivals who bid the symmetric first-price equilibrium (\emph{FP eq.}). We fit a lognormal value distribution $F$ per product by maximum likelihood, treating each closing price as the second-highest of $n$ bidders' values, and let $N=n+1$ rivals bid $\beta(x)=x-\int_0^x F(t)^{N-1}\,dt\,/\,F(x)^{N-1}$, so that $b$ wins with probability $F(\beta^{-1}(b))^{n}$ averaged over auctions. The calculator's bid matches the FP-optimal bid in 58\% of scenarios.

\section{Models and protocol}

\subsection{Models}
The checkpoints are Phi-4-mini-reasoning (Phi-mini) \cite{microsoft2025phireasoning}, Qwen3-4B, -8B and -14B with thinking enabled \cite{yang2025qwen3}, DeepSeek-R1-Distill-Llama-8B (DeepSeek-8B) \cite{guo2025deepseek} and Granite-3.2-8B-Instruct (Granite-8B) with reasoning on \cite{ibmgranite2025}.

\subsection{Prompt and forcing}
The prompt (Appendix~\ref{app:prompts}) gives the private value, lists the options in randomized order with their bids and estimated win probabilities, states the objective as $(v-b)\times\hat p(b)$, and asks the model not to infer the answer from option position. For each budget $c\in\{0,64,128,256,512,768\}$ we force exactly $c$ reasoning tokens by suppressing end-of-sequence until $c$ tokens are generated; the end-of-thinking marker itself is not suppressed. We then append a readout prompt and take the most likely option letter among the menu's letters (constrained readout). Decoding is greedy, and each scenario is run under three option orders (seeds 11, 23 and 37), giving 648 decisions per model and budget. Greedy forced traces are prefixes of each other across budgets, so we can watch the same trace grow.

Forcing is a measurement device. The deployment-relevant number is the unforced run, in which the model stops on its own within a cap of 1{,}024 tokens and, for all models but Granite-8B, within 4{,}096 tokens.

\subsection{Readout}
The paper's readout, which we also call the open-block readout, appends ``Use the reasoning above and select the best displayed option. Return one option letter only. FINAL OPTION:'' directly after the reasoning tokens, without closing the reasoning block. In the closed-block ablation (Section~\ref{sec:readout}) we first append the model's own end-of-thinking marker (\texttt{</think>} for the five \texttt{<think>} models, Granite's ``Here is my response:'' header for Granite-8B), then ``Final answer: Option'', and read the most likely letter. Gap recovered under the closed-block readout keeps the open-block $q(0)$ as its baseline.

\subsection{Hardware, cost and sampling}
The main runs used HF transformers with batch size 1 on an RTX 4090 (an L40S for Qwen3-14B). The 768 tokens of one decision cost $c_{768}=0.19$--$0.75$\textcent{} at measured unbatched speed and RunPod rates of \$0.74--0.79 per GPU-hour \cite{runpod2026pricing}; batched serving would cost less. The \emph{break-even} $c_{768}/(q(768)-q(0))$ is the best expected profit per decision above which 768 tokens pay for themselves. Five sampled seeds (temperature 0.6, top-$p$ 0.95) at 128, 384 and 768 tokens move $q$ by at most 0.017. The closed-block ablation and all AuctionNet runs used vLLM on an L40S; on eBay this setup moves Phi-mini's and Granite-8B's $q$ by at most 0.010 at 768 tokens and 0.027 without forcing, and it reproduces 94\% of Phi-mini's saved letters and 99--100\% of the other models'.

\section{Reading the traces}
\label{sec:parsing}

Two pieces of code read each trace; no human annotator is involved.
\begin{itemize}
\item The \emph{text reader} splits the trace at each mention of an option and takes the last result number in that segment, within a fixed character window, as the trace's value for that option.
\item The \emph{matcher} marks an option as evaluated when the trace states its true expected profit within 2\%.
\end{itemize}
An option's \emph{written value} is its true value when the matcher finds it and the reader's number otherwise. This bounds arithmetic errors from below; counting every misread value as miscomputed bounds them from above. Where both pieces of code find a value, the reader's number equals the true value in 86--95\% of options. A trace is \emph{finished} when the matcher finds every option's value.

Each wrong 768-token answer falls in the first category that applies:
\begin{description}
\item[Not reached] The trace never writes the best option's value.
\item[Arithmetic] The trace's own written values rank another option above the best, so a profit was miscomputed.
\item[Cut off] The best option is written and ranked first, the list is unfinished, and the answer is another bid.
\item[Passed over] The best option is written and ranked first, the list is finished, and the answer is another bid.
\end{description}
For example, in one Qwen3-4B trace the calculator's bid A is computed and ranked first, the trace is cut off while computing option E, and the model answers C, the highest bid; without forcing, the same trace finishes E and picks A.

A separate rule-based parser reads the \emph{stated answer}: the last explicit choice the trace itself states (``the highest expected profit is Option C''). We use it to separate readout errors from reasoning errors.

\section{Results on eBay}

\begin{figure*}[t]
\centering
\includegraphics[width=\textwidth]{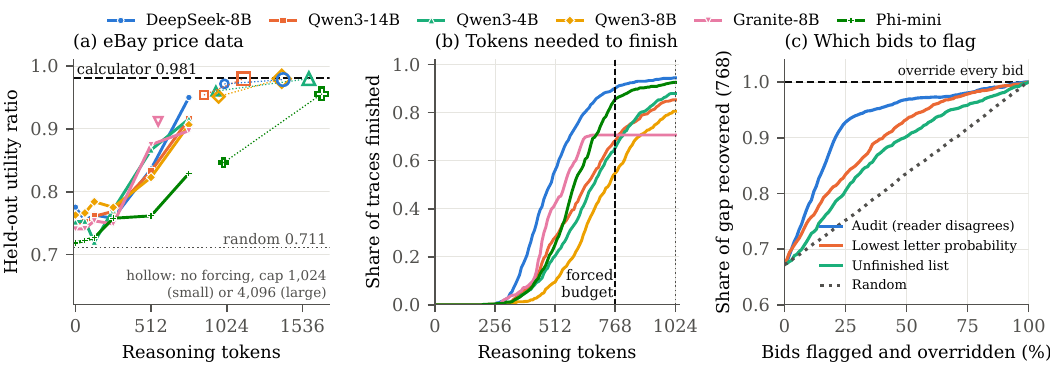}
\caption{eBay. (a) Held-out utility ratio by forced budget (filled markers, 648 decisions per point) and without forcing at mean token use (hollow; small: 1{,}024 cap, large: 4{,}096 cap). (b) Share of unforced traces that have evaluated every option by token $t$. (c) Share of the 768-token gap recovered when a given share of bids is flagged and overridden with the calculator's bid, by flagging rule, mean over the six models; the audit first flags bids whose answer's written value falls furthest below the largest.}
\label{fig:main}
\Description{Three line plots: utility ratio against reasoning tokens for six models; share of unforced traces that have evaluated every option against tokens; share of the gap recovered against the share of bids flagged for override, for four flagging rules.}
\end{figure*}

\begin{table*}[t]
\caption{Held-out utility ratio $q$ on eBay by forced reasoning budget, and without forcing under a 1,024-token and a 4,096-token cap. Calculator: 0.981; random choice: 0.711; 648 decisions per cell.}
\label{tab:qbudget}
\small
\centering

\begin{tabular}{lcccccccc}
\toprule
Model & $q(0)$ & $q(64)$ & $q(128)$ & $q(256)$ & $q(512)$ & $q(768)$ & 1,024 run & 4,096 run \\
\midrule
DeepSeek-8B & 0.775 & 0.754 & 0.763 & 0.758 & 0.836 & 0.950 & 0.972 & 0.978 \\
Phi-mini & 0.718 & 0.722 & 0.727 & 0.758 & 0.762 & 0.829 & 0.846 & 0.956 \\
Qwen3-4B & 0.752 & 0.753 & 0.720 & 0.771 & 0.867 & 0.916 & 0.960 & 0.980 \\
Qwen3-8B & 0.763 & 0.766 & 0.784 & 0.775 & 0.823 & 0.907 & 0.951 & 0.980 \\
Qwen3-14B & 0.749 & 0.751 & 0.761 & 0.769 & 0.833 & 0.916 & 0.954 & 0.980 \\
Granite-8B & 0.740 & 0.741 & 0.754 & 0.749 & 0.875 & 0.897 & 0.912 & -- \\
\bottomrule
\end{tabular}
\end{table*}

\begin{table*}[t]
\caption{Gap to the calculator recovered on eBay, \% [95\% family-bootstrap interval]. Columns 256--768: forced budgets with the paper's letter readout. Stated: the answer the trace states, where it states one, and the letter otherwise. 1,024 and 4,096 run: no forcing, letter read after the model stops. FP eq.: 768 tokens scored against the first-price equilibrium win curve.}
\label{tab:gapebay}
\footnotesize
\centering
\setlength{\tabcolsep}{4pt}
\begin{tabular}{lccccccc}
\toprule
Model & 256 & 512 & 768 & 768, stated & 1,024 run & 4,096 run & 768, FP eq. \\
\midrule
DeepSeek-8B & $-$9 [$-$17, 0] & 30 [22, 37] & 85 [78, 92] & 77 [69, 85] & 96 [93, 98] & 99 [98, 100] & 85 [77, 92] \\
Phi-mini & 15 [7, 22] & 17 [8, 25] & 42 [34, 50] & 64 [56, 71] & 49 [40, 57] & 91 [87, 94] & 40 [28, 49] \\
Qwen3-4B & 8 [$-$3, 19] & 50 [43, 58] & 72 [66, 78] & 73 [66, 80] & 91 [87, 96] & 100 [100, 100] & 74 [69, 80] \\
Qwen3-8B & 6 [$-$1, 12] & 28 [22, 34] & 66 [60, 73] & 66 [58, 74] & 87 [82, 92] & 100 [100, 100] & 67 [61, 73] \\
Qwen3-14B & 8 [3, 14] & 36 [29, 43] & 72 [66, 78] & 73 [66, 80] & 89 [83, 94] & 100 [100, 100] & 75 [68, 81] \\
Granite-8B & 3 [$-$2, 9] & 56 [47, 64] & 65 [57, 73] & 72 [66, 77] & 72 [66, 77] & -- & 74 [68, 80] \\
\bottomrule
\end{tabular}
\end{table*}

\subsection{Overview}
Table~\ref{tab:qbudget} gives the utility ratio at every budget and Table~\ref{tab:gapebay} the gap recovered. Below 256 tokens reasoning buys almost nothing, and DeepSeek-8B is slightly worse at 256 tokens than with none. Between 512 and 768 tokens most models recover most of what they will recover under forcing: at 768 tokens the six models recover 42--85\% of the calculator's lead, and in dollars pooled over decisions 25--75\%. Without forcing, under a 1{,}024-token cap, they recover 49--96\%, and the five models we also ran with a 4{,}096-token cap recover 91--100\%.

Table~\ref{tab:splitebay} shows where the 768-token gap goes. For every model most of the loss comes after the best option's value is on the page. For five models arithmetic is at most 5\% of the gap with the lower bound and at most 17\% with the upper bound; Granite-8B is the exception. The shares move by at most 11 points when the matcher's tolerance is 1\% or 5\% instead of 2\%, or when bids are scored against the first-price equilibrium curve.

\begin{table*}[t]
\caption{Where the eBay 768-token gap goes, \% of the gap [95\% interval], letters read with the block open; rows sum to 100 up to rounding, and intervals can dip below zero because a wrong answer can beat the calculator on held-out data. Finished: forced 768-token traces in which the matcher finds every option's value. Median: tokens by which half of unforced traces finish ($K=5$/$K=7$). Arithmetic uses the lower bound (matcher value where found); Arith., upper counts every misread value as miscomputed.}
\label{tab:splitebay}
\footnotesize
\centering

\begin{tabular}{lcccccc c}
\toprule
Model & Finished & Median & Not reached & Arithmetic & Cut off & Passed over & Arith., upper \\
\midrule
DeepSeek-8B & 90 & 440/544 & 2 [0, 7] & 5 [0, 17] & 44 [24, 61] & 48 [32, 67] & 7 \\
Phi-mini & 84 & 544/664 & 5 [2, 9] & 2 [0, 3] & 11 [7, 17] & 82 [74, 89] & 17 \\
Qwen3-4B & 64 & 608/776 & 15 [8, 24] & 0 [0, 0] & 63 [51, 73] & 22 [12, 34] & 3 \\
Qwen3-8B & 53 & 656/840 & 34 [24, 44] & 4 [0, 8] & 39 [29, 50] & 23 [15, 32] & 12 \\
Qwen3-14B & 66 & 616/704 & 20 [11, 30] & 3 [$-$1, 9] & 59 [49, 70] & 18 [7, 30] & 3 \\
Granite-8B & 71 & 464/616 & 11 [6, 17] & 22 [13, 32] & 6 [1, 11] & 61 [51, 70] & 45 \\
\bottomrule
\end{tabular}
\end{table*}

\subsection{Budget: the Qwen models and DeepSeek-8B}
These four models lose mainly to the token cap. Their gap is mostly \emph{not reached} or \emph{cut off} (Table~\ref{tab:splitebay}), and on 7-option menus half of their unforced traces have evaluated every option only after 544--840 tokens. Because greedy forced traces are prefixes of each other, we can watch a list finish between 512 and 768 tokens: when it does, these models answer correctly in 84--89\% of cases. Allowed to stop on their own within 4{,}096 tokens, they rarely hit the cap (at most 1.1\% of decisions), stop after a median of 980--1{,}390 tokens, and recover 99--100\% of the gap (Figure~\ref{fig:main}a, Table~\ref{tab:gapebay}). Under a 1{,}024-token cap the Qwen models are still often cut off: cut-off lists make up 46--52\% of their remaining gap (Table~\ref{tab:splitebay2}), which is why the larger cap closes it.

The extra tokens are cheap relative to the stakes. Those tokens cost 0.45--1.11\textcent{} per decision, while the smallest best expected profit in any scenario (\$1.75) is 39--130 times the 768-token break-even and the median is 1{,}076--3{,}608 times it (Table~\ref{tab:cost}). For these models the fix is not to cap reasoning.

Asking for the list in the prompt does not help either (Table~\ref{tab:prompt}). One added instruction (``End by listing the expected profit of each option, then answer with the option that has the largest.'') changes recovery by at most 5 points for Phi-mini, Granite-8B and DeepSeek-8B, with intervals that include zero, and lowers it by 22 points [14, 31] for Qwen3-8B, which stood in for the Qwen family: its longer lists no longer fit, and 61 of its 107 new errors are lists cut off before the end. Without forcing it still loses 14 points [8, 21].

\begin{table}[t]
\caption{Cost of 768 reasoning tokens per decision at measured unbatched speed and \$0.74--0.79 per GPU-hour, and the break-even best expected profit $c_{768}/(q(768)-q(0))$. Min.\ and Median: the smallest and median best expected profit across scenarios, as multiples of the break-even. Gain / cost: mean dollar gain from the 768 tokens over their cost.}
\label{tab:cost}
\small
\centering

\begin{tabular}{lccccc}
\toprule
Model & Cost (\textcent) & Break-even (\textcent) & Min.\ / B-e & Median / B-e & Gain / cost \\
\midrule
DeepSeek-8B & 0.31 & 1.78 & 98 & 2,718 & 9,536 \\
Phi-mini & 0.19 & 1.71 & 102 & 2,834 & 5,872 \\
Qwen3-4B & 0.22 & 1.34 & 130 & 3,608 & 13,881 \\
Qwen3-8B & 0.34 & 2.34 & 75 & 2,065 & 9,263 \\
Qwen3-14B & 0.75 & 4.49 & 39 & 1,076 & 4,223 \\
Granite-8B & 0.34 & 2.19 & 80 & 2,208 & 13,527 \\
\bottomrule
\end{tabular}
\end{table}

\begin{table}[t]
\caption{Adding one instruction (``End by listing the expected profit of each option, then answer with the option that has the largest.''): gap recovered, \% [95\% interval]. Original prompt: as in Table~\ref{tab:gapebay}.}
\label{tab:prompt}
\footnotesize
\centering

\begin{tabular}{llccc}
\toprule
Model & Setting & Original prompt & With list instruction & Change \\
\midrule
Phi-mini & 768 & 42 [34, 50] & 38 [29, 45] & $-$5 [$-$13, 4] \\
Phi-mini & 1,024 run & 49 [40, 57] & 54 [43, 62] & 5 [$-$5, 13] \\
Granite-8B & 768 & 65 [57, 73] & 68 [60, 76] & 3 [$-$4, 10] \\
Granite-8B & 1,024 run & 72 [66, 77] & 77 [70, 83] & 5 [$-$3, 13] \\
DeepSeek-8B & 768 & 85 [78, 92] & 82 [75, 89] & $-$3 [$-$8, 2] \\
DeepSeek-8B & 1,024 run & 96 [93, 98] & 94 [90, 97] & $-$2 [$-$6, 2] \\
Qwen3-8B & 768 & 66 [60, 73] & 44 [36, 52] & $-$22 [$-$31, $-$14] \\
Qwen3-8B & 1,024 run & 87 [82, 92] & 73 [66, 80] & $-$14 [$-$21, $-$8] \\
\bottomrule
\end{tabular}
\end{table}

\subsection{Readout: Phi-mini}
\label{sec:readout}
Phi-mini's loss is mostly a readout effect. Its trace usually states the right option, but its letter departs from it. The stated-answer parser agrees with the letter in 96--98\% of Qwen traces but in only 57\% of Phi-mini's; where Phi-mini's trace states a choice, that choice is the best option 94\% of the time, while its letter is the best option 55\% of the time. Scoring the stated answer where there is one raises Phi-mini's gap recovered from 42\% to 64\% (Table~\ref{tab:gapebay}).

The cause is the readout prompt, which asks for the letter while the reasoning block is still open. When we first close the block with the model's own end-of-thinking marker, Phi-mini's letter agrees with its stated answer in 95\% of decisions and recovers 84\% of the gap (Table~\ref{tab:readout}). The same change adds 5--14 points for DeepSeek-8B, Granite-8B, Qwen3-8B and Qwen3-14B, with intervals above zero, and nothing for Qwen3-4B. Left to close the block itself within 4{,}096 tokens, Phi-mini recovers 91\% [87, 94], with a median of 1{,}364 tokens and 6.8\% of decisions at the cap. Table~\ref{tab:splitebay2} gives the split under the closed readout: for the Qwen models, not reached and cut off then make up 81--93\% of what remains.

\begin{table*}[t]
\caption{Closing the reasoning block before reading the answer, at 768 forced tokens. Repro.: share of the saved letters that the rerun reproduces before the change (\%). Agree: share of decisions where the letter equals the answer the trace states (\%). Open, Closed, Change: gap recovered with the paper's readout (rerun letters), with the closed-block readout, and their difference, \% [95\% interval].}
\label{tab:readout}
\scriptsize
\centering
\setlength{\tabcolsep}{3pt}
\begin{tabular}{lccccccccc}
\toprule
& \multicolumn{6}{c}{eBay} & \multicolumn{3}{c}{AuctionNet} \\ \cmidrule(lr){2-7}\cmidrule(lr){8-10} Model & Repro. & Agree, open & Agree, closed & Open & Closed & Change & Open & Closed & Change \\
\midrule
DeepSeek-8B & 99 & 89 & 89 & 85 [77, 91] & 91 [86, 95] & 6 [1, 11] & 92 [89, 95] & 95 [92, 97] & 2 [1, 4] \\
Phi-mini & 94 & 57 & 95 & 41 [31, 50] & 84 [79, 88] & 43 [33, 54] & 24 [15, 32] & 81 [75, 86] & 57 [48, 66] \\
Qwen3-4B & 99 & 96 & 97 & 72 [65, 79] & 72 [63, 79] & 0 [$-$6, 5] & 80 [75, 85] & 81 [76, 87] & 2 [$-$2, 6] \\
Qwen3-8B & 99 & 98 & 99 & 67 [60, 75] & 77 [70, 85] & 10 [4, 16] & 67 [62, 73] & 82 [78, 86] & 15 [11, 19] \\
Qwen3-14B & 100 & 98 & 98 & 72 [65, 79] & 86 [80, 92] & 14 [10, 19] & 83 [78, 87] & 94 [92, 96] & 11 [8, 15] \\
Granite-8B & 99 & 91 & 96 & 65 [58, 72] & 70 [63, 76] & 5 [2, 9] & 73 [68, 78] & 76 [72, 81] & 3 [1, 5] \\
\bottomrule
\end{tabular}
\end{table*}

\begin{table*}[t]
\caption{The eBay gap split under the closed-block readout (\% [95\% interval]) and for the unforced 1,024-token run (\%, point estimates). NR: not reached; Ar: arithmetic (lower bound); Cut: cut off; Pass: passed over.}
\label{tab:splitebay2}
\footnotesize
\centering

\begin{tabular}{lcccc cccc}
\toprule
& \multicolumn{4}{c}{768 tokens, closed block} & \multicolumn{4}{c}{No forcing, 1,024 cap} \\ \cmidrule(lr){2-5}\cmidrule(lr){6-9} Model & NR & Ar & Cut & Pass & NR & Ar & Cut & Pass \\
\midrule
DeepSeek-8B & 4 [0, 12] & 8 [0, 21] & 46 [22, 71] & 42 [23, 62] & 6 & 0 & 0 & 94 \\
Phi-mini & 8 [2, 17] & 3 [0, 8] & 27 [15, 39] & 62 [50, 74] & 1 & 1 & 2 & 97 \\
Qwen3-4B & 19 [10, 27] & 0 [$-$1, 1] & 62 [51, 74] & 19 [8, 32] & 9 & 1 & 46 & 44 \\
Qwen3-8B & 37 [26, 48] & 4 [0, 10] & 52 [42, 62] & 7 [$-$3, 18] & 30 & $-$1 & 52 & 20 \\
Qwen3-14B & 32 [15, 54] & 2 [$-$8, 9] & 61 [44, 77] & 5 [$-$10, 20] & 13 & 3 & 49 & 36 \\
Granite-8B & 13 [7, 19] & 25 [15, 37] & 4 [0, 9] & 58 [47, 68] & 14 & 27 & 4 & 55 \\
\bottomrule
\end{tabular}
\end{table*}

\subsection{Decision and arithmetic: Granite-8B}
Granite-8B is the one model whose loss is partly arithmetic: 22\% of its gap with the lower bound and 45\% with the upper bound, against 61\% and 37\% passed over. It is not a readout effect: its stated answer agrees with its letter in 93\% of finished traces, and with the block closed 58\% of its gap is still passed over. Nor is it caused by forcing. Granite-8B stops on its own before 768 tokens in almost every decision, and on its unforced traces passing over is still 55\% of the gap.

\subsection{Which wrong bids, and why}
\label{sec:wrong}
Wrong answers lean toward the highest bid on the menu, the one most likely to win (Table~\ref{tab:wrong}). For DeepSeek-8B and the Qwen models the highest bid accounts for 40--48\% of wrong answers, against about 18\% if a wrong answer were a random non-calculator option.

\paragraph{Size, not position}
Option orders are randomized, so a bid's size and its place in the list are independent. A conditional logit over the non-calculator options of each wrong answer (Table~\ref{tab:possize}) shows that, with display position held fixed, a wrong answer from DeepSeek-8B or a Qwen model is far more likely to be the highest bid than any other wrong option (odds ratio 3.9 [2.9, 5.1]), while being listed first or last makes an option less likely, not more (odds ratios 0.59 and 0.50). The size effect is the same on AuctionNet (3.9 [3.2, 4.8]). Across all six models the size effect is smaller but clear (odds ratios 2.3 on eBay and 2.2 on AuctionNet), and on AuctionNet being listed first raises the odds somewhat (1.38 [1.20, 1.57]).

\paragraph{Not a protocol artifact}
The lean survives the closed-block readout (38--45\% of wrong answers for these four models) and sampling at temperature 0.6 over five seeds (40--52\%).

\paragraph{Not a difficulty effect}
A logistic regression for passing over (Table~\ref{tab:reg}), on the 3{,}519 traces that write the best option's value and rank it first, whether or not the list is finished, holds the relative profit gap between the best and runner-up option, the best option's position, $K$, value, product and model fixed. The odds of passing over are about a third as large when the best option is itself the highest bid (odds ratio 0.36 [0.28, 0.46]), under half as large when the list is finished (0.38), and nearly twice as large on 7-option menus (1.89). A mixed logit with a scenario random intercept gives the same odds ratios. Per-model slopes are in Appendix~\ref{app:extra} (Table~\ref{tab:regmodel}): the highest-bid effect is strongest for Qwen3-8B and Qwen3-14B, and position matters in opposite directions for Phi-mini (later is worse) and Granite-8B (later is better).

\paragraph{What it costs}
At 768 forced tokens and under pay-your-bid pricing these choices cost bidders 7--17\% of the surplus the calculator's bids earn, or \$15k--\$36k per 1{,}000 decisions on these products (\$10k--\$38k with the closed-block readout). For DeepSeek-8B and the Qwen models, 51--78\% of that loss comes from choosing the highest bid. Against rivals in first-price equilibrium the calculator's bids would raise bidder surplus by \$12k--\$28k per 1{,}000 decisions. Under eBay's own rule, where the winner pays about the second-highest bid, higher bids do not hurt the bidder, which is why we frame the cost for pay-your-bid settings.

\begin{table*}[t]
\caption{Wrong answers at 768 tokens and their cost on eBay. Wrong: number of the 648 answers that differ from the calculator. Highest: share of wrong answers that are the menu's highest bid (\%), against the share expected if a wrong answer were a uniformly random non-calculator option (Chance). Closed: same share with the closed-block readout. Sampled: temperature 0.6, top-$p$ 0.95, five seeds. AuctionNet: same share on the second market. Loss: pay-your-bid surplus lost relative to the calculator's bids, as \% of the calculator's surplus and in thousands of dollars per 1,000 decisions. From highest: share of that loss due to answers that are the highest bid (\%).}
\label{tab:wrong}
\footnotesize
\centering

\begin{tabular}{lccccccccc}
\toprule
Model & Wrong & Highest & Chance & Closed & Sampled & AuctionNet & Loss & \$k/1{,}000 & From highest \\
\midrule
DeepSeek-8B & 89 & 44 & 19 & 38 & 43 & 21 & 11 & 22.6 & 78 \\
Phi-mini & 375 & 22 & 17 & 42 & 23 & 23 & 16 & 33.7 & 37 \\
Qwen3-4B & 187 & 40 & 18 & 45 & 40 & 33 & 14 & 30.5 & 60 \\
Qwen3-8B & 209 & 48 & 18 & 42 & 51 & 52 & 14 & 30.4 & 51 \\
Qwen3-14B & 151 & 48 & 18 & 38 & 52 & 58 & 17 & 36.0 & 68 \\
Granite-8B & 301 & 20 & 17 & 21 & 19 & 11 & 7 & 15.1 & 26 \\
\bottomrule
\end{tabular}
\end{table*}

\begin{table}[t]
\caption{Size or position? Conditional logit over the non-calculator options of each wrong 768-token answer, odds ratios [95\% scenario-bootstrap interval]. Option order is randomized, so position is independent of size.}
\label{tab:possize}
\footnotesize
\centering

\begin{tabular}{lcccc}
\toprule
Sample & Wrong answers & Highest bid & Listed first & Listed last \\
\midrule
eBay, DeepSeek-8B and Qwen & 636 & 3.91 [2.93, 5.13] & 0.59 [0.34, 0.93] & 0.50 [0.32, 0.70] \\
eBay, all models & 1312 & 2.31 [1.93, 2.78] & 1.10 [0.88, 1.32] & 0.54 [0.43, 0.67] \\
AuctionNet, DeepSeek-8B and Qwen & 619 & 3.91 [3.22, 4.84] & 0.40 [0.25, 0.58] & 0.96 [0.70, 1.20] \\
AuctionNet, all models & 1360 & 2.17 [1.87, 2.55] & 1.38 [1.20, 1.57] & 0.87 [0.69, 1.01] \\
\bottomrule
\end{tabular}
\end{table}

\begin{table}[t]
\caption{Passing over the best option: logistic regression, odds ratios [95\% interval], errors clustered by scenario. Sample: 768-token traces that write the best option's value and rank it first (eBay $n=3519$, 1074 passed over; AuctionNet $n=3614$, 1133 passed over). Product (category) and model fixed effects are included; the reference model is Qwen3-8B.}
\label{tab:reg}
\small
\centering

\begin{tabular}{lcc}
\toprule
Covariate & eBay & AuctionNet \\
\midrule
Best option is the highest bid & 0.36 [0.28, 0.46] & 0.29 [0.22, 0.38] \\
List finished & 0.38 [0.29, 0.51] & 0.43 [0.32, 0.58] \\
Relative profit gap, best vs runner-up & 0.36 [0.13, 0.99] & 0.09 [0.03, 0.26] \\
Position of best option (0 first, 1 last) & 1.93 [1.42, 2.61] & 2.48 [1.84, 3.35] \\
$K=7$ & 1.89 [1.48, 2.41] & 1.95 [1.57, 2.43] \\
log value relative to product median & 1.08 [0.77, 1.51] & 3.10 [1.98, 4.83] \\
\bottomrule
\end{tabular}
\end{table}

\subsection{The audit}
\label{sec:audit}
Checking the answer against the trace's own numbers catches most errors. The text reader alone does not replace the answer: its gains are in-sample, and on synthetic traces it hurts four models. As a flag it works. When the reader's choice from the trace's written values disagrees with the answer, we flag the bid; in our experiments a flagged bid is overridden with the calculator's bid. The flag itself uses only the trace, so an operator needs no calculator to decide which bids to review.

Table~\ref{tab:audit} gives gap recovered under several out-of-sample designs. With the reader's settings (window and number of required values) chosen on synthetic traces and applied to eBay, the audit recovers 84--100\% of the gap, against 42--85\% for the answer alone; under nested cross-validation with settings chosen on two products and tested on the third, or on five of six family folds, it recovers the same. A flagged answer is wrong in 58--89\% of cases, and the audit catches 53--88\% of wrong answers. At every flag rate it beats flagging by low letter probability, by unfinished list, or at random (Figure~\ref{fig:main}c, Table~\ref{tab:flagrate}): flagging 20\% of bids recovers 89\% of the gap, against at most 81\% for the other rules. The gain is not specific to the open-block readout: with the closed-block letters, where the answer alone recovers 70--91\%, the same nested audit recovers 84--100\% while flagging 15--33\% of bids, and flagging 20\% recovers 95\% against 84\% at random. We do not price the review itself.

\begin{table*}[t]
\caption{The audit: gap recovered (\%) when bids on which the text reader and the answer disagree are flagged and overridden with the calculator's bid. Answer: no audit. Syn.$\to$eBay: reader setting chosen on synthetic traces. LPO: nested, setting chosen on two products and tested on the third. Fam.: nested, family 6-fold. eBay$\to$Syn.: chosen on eBay, tested on synthetic traces. Flags: share of decisions flagged (\%). AuctionNet: setting chosen on eBay alone, and nested leave-one-category-out.}
\label{tab:audit}
\footnotesize
\centering

\begin{tabular}{lccccccccc}
\toprule
& \multicolumn{6}{c}{eBay} & \multicolumn{3}{c}{AuctionNet} \\ \cmidrule(lr){2-7}\cmidrule(lr){8-10} Model & Answer & Syn.$\to$eBay & LPO & Fam. & eBay$\to$Syn. & Flags & eBay set. & LOCO & Flags \\
\midrule
DeepSeek-8B & 85 & 100 & 100 & 100 & 98 & 18 & 99 & 99 & 15 \\
Phi-mini & 42 & 98 & 98 & 97 & 95 & 57 & 98 & 98 & 63 \\
Qwen3-4B & 72 & 98 & 98 & 98 & 98 & 33 & 99 & 99 & 29 \\
Qwen3-8B & 66 & 96 & 96 & 96 & 99 & 33 & 96 & 96 & 36 \\
Qwen3-14B & 72 & 99 & 99 & 99 & 100 & 32 & 99 & 99 & 31 \\
Granite-8B & 65 & 84 & 84 & 84 & 81 & 34 & 86 & 86 & 29 \\
\bottomrule
\end{tabular}
\end{table*}

\begin{table}[t]
\caption{Gap recovered (\%, mean over models) when a fixed share of bids is flagged and overridden, by flagging rule.}
\label{tab:flagrate}
\small
\centering

\begin{tabular}{lcccc}
\toprule
Flag & 10\% & 20\% & 30\% & AuctionNet, 20\% \\
\midrule
Audit (reader disagrees) & 78 & 89 & 94 & 88 \\
Lowest letter probability & 75 & 81 & 85 & 81 \\
Unfinished list & 72 & 78 & 83 & 79 \\
Random & 70 & 74 & 77 & 76 \\
\bottomrule
\end{tabular}
\end{table}

\section{Results on AuctionNet}
\label{sec:anresults}

\begin{figure*}[t]
\centering
\includegraphics[width=0.9\textwidth]{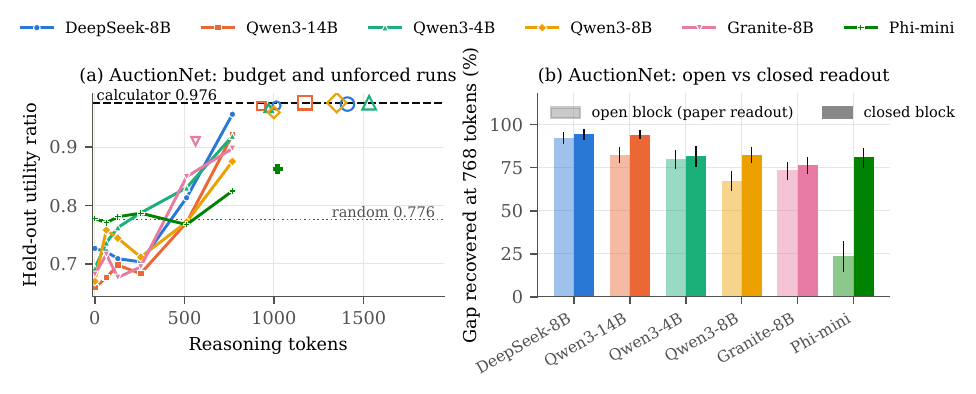}
\caption{AuctionNet. (a) Held-out utility ratio by forced budget, and without forcing at mean token use (hollow; small: 1{,}024 cap, large: 4{,}096 cap). (b) Gap recovered at 768 forced tokens with the paper's readout (light) and the closed-block readout (dark), with 95\% intervals.}
\label{fig:an}
\Description{Left: utility ratio against reasoning tokens for six models on AuctionNet, with hollow markers for unforced runs near the calculator line. Right: paired bars of gap recovered with open and closed readout per model; Phi-mini rises from about a quarter to about four fifths.}
\end{figure*}

\begin{table*}[t]
\caption{AuctionNet: utility ratio and gap recovered, \% [95\% interval; families resampled within category]. Calculator $q$ = 0.976; random choice 0.776. Dollars: pooled over decisions instead of averaged per decision.}
\label{tab:anmain}
\footnotesize
\centering

\begin{tabular}{lccccccc}
\toprule
Model & $q(0)$ & $q(768)$ & 768 & 768, closed & 1,024 run & 4,096 run & 768, dollars \\
\midrule
DeepSeek-8B & 0.726 & 0.956 & 92 [89, 95] & 95 [92, 97] & 98 [96, 99] & 99 [98, 100] & 89 \\
Phi-mini & 0.778 & 0.825 & 24 [15, 32] & 81 [75, 86] & 43 [35, 51] & -- & 26 \\
Qwen3-4B & 0.692 & 0.919 & 80 [75, 85] & 81 [76, 87] & 97 [95, 98] & 100 [100, 100] & 74 \\
Qwen3-8B & 0.670 & 0.875 & 67 [62, 73] & 82 [78, 86] & 95 [92, 97] & 100 [100, 100] & 62 \\
Qwen3-14B & 0.659 & 0.921 & 83 [78, 87] & 94 [92, 96] & 98 [97, 100] & 100 [100, 100] & 76 \\
Granite-8B & 0.682 & 0.898 & 73 [68, 78] & 76 [72, 81] & 78 [74, 82] & -- & 71 \\
\bottomrule
\end{tabular}
\end{table*}

\begin{table*}[t]
\caption{Where the AuctionNet 768-token gap goes, \% of the gap. Open block: [95\% interval]; closed block: point estimates. Fin.: forced traces that evaluate every option. Abbreviations as in Table~\ref{tab:splitebay2}.}
\label{tab:ansplit}
\footnotesize
\centering

\begin{tabular}{lc cccc cccc}
\toprule
& & \multicolumn{4}{c}{Open block} & \multicolumn{4}{c}{Closed block} \\ \cmidrule(lr){3-6}\cmidrule(lr){7-10} Model & Fin. & NR & Ar & Cut & Pass & NR & Ar & Cut & Pass \\
\midrule
DeepSeek-8B & 88 & 8 [2, 18] & 0 [0, 0] & 27 [10, 43] & 65 [48, 82] & 8 & 0 & 43 & 49 \\
Phi-mini & 93 & 3 [1, 6] & 1 [0, 3] & 6 [3, 9] & 90 [86, 94] & 14 & 2 & 14 & 70 \\
Qwen3-4B & 75 & 22 [13, 33] & 1 [0, 3] & 42 [28, 55] & 35 [23, 48] & 26 & 1 & 47 & 26 \\
Qwen3-8B & 58 & 22 [15, 28] & 3 [0, 6] & 55 [45, 64] & 21 [14, 29] & 27 & 4 & 56 & 13 \\
Qwen3-14B & 81 & 11 [5, 17] & 2 [0, 7] & 48 [35, 59] & 39 [29, 51] & 22 & 2 & 54 & 22 \\
Granite-8B & 91 & 9 [5, 15] & 19 [13, 25] & 5 [1, 10] & 67 [59, 74] & 11 & 20 & 5 & 64 \\
\bottomrule
\end{tabular}
\end{table*}

The pattern holds on the second market (Figure~\ref{fig:an}, Tables~\ref{tab:anmain} and~\ref{tab:ansplit}).
\begin{itemize}
\item \textbf{Budget.} At 768 forced tokens the models recover 24--92\% of the gap. Without forcing, DeepSeek-8B and the Qwen models recover 95--98\% within 1{,}024 tokens and 99--100\% within 4{,}096, hitting the larger cap in at most 0.9\% of decisions after a median of 1{,}047--1{,}474 tokens. With no reasoning, the models score 0.659--0.778, mostly below random choice (0.776), so short reasoning starts from a low base.
\item \textbf{Readout.} Closing the reasoning block lifts Phi-mini from 24\% to 81\% [75, 86] and moves the others by 2--15 points; Qwen3-4B's change is again indistinguishable from zero (Table~\ref{tab:readout}).
\item \textbf{Where the gap goes.} Arithmetic is at most 3\% of the gap for five models and 19\% for Granite-8B. Not reached and cut off make up 59--77\% of the Qwen models' gap, and passed over makes up 90\% of Phi-mini's with the open readout and 70\% with the closed one.
\item \textbf{Which wrong bids.} The highest bid is 21--58\% of wrong answers for DeepSeek-8B and the Qwen models (Table~\ref{tab:wrong}), weaker for DeepSeek-8B than on eBay. The size-not-position result replicates (Table~\ref{tab:possize}), and in the same regression passing over is again much less likely when the best option is the highest bid (odds ratio 0.29 [0.22, 0.38]; Table~\ref{tab:reg}).
\item \textbf{Audit.} With the reader setting chosen on eBay alone, the audit recovers 86--99\%, and the same holds with nested leave-one-category-out selection. Flagging 20\% of bids recovers 88\%, against at most 81\% for the other rules (Tables~\ref{tab:audit} and~\ref{tab:flagrate}).
\end{itemize}
The parser was checked for the different scale of AuctionNet: with a strictly relative 2\% matching tolerance, the split is unchanged.

\section{Robustness}
\label{sec:robust}

\begin{table*}[t]
\caption{Robustness of gap recovered at 768 tokens (\%). Family boot.: the main interval. Product boot.: resampling the three products, then families within them. Calc. ahead: share of landscape-bootstrap draws (construction and held-out auctions both resampled) in which the calculator beats the model (\%). First-price variants: value distribution (lognormal, gamma, Weibull), a coarser bid increment, and one bidder fewer or more. Logn.\ refits the curve behind the FP eq.\ column of Table~\ref{tab:gapebay}; the two differ by under one point.}
\label{tab:robust}
\footnotesize
\centering

\begin{tabular}{lccc cccccc}
\toprule
& \multicolumn{3}{c}{Empirical curve} & \multicolumn{6}{c}{First-price equilibrium curve, variants} \\ \cmidrule(lr){2-4}\cmidrule(lr){5-10} Model & Family boot. & Product boot. & Calc. ahead & Logn. & Gamma & Weib. & Incr. & $n-1$ & $n+1$ \\
\midrule
DeepSeek-8B & 85 [78, 92] & 85 [67, 99] & 96.3 & 85 & 86 & 86 & 85 & 84 & 86 \\
Phi-mini & 42 [34, 50] & 42 [10, 59] & 100.0 & 39 & 39 & 38 & 39 & 36 & 43 \\
Qwen3-4B & 72 [66, 78] & 72 [54, 87] & 99.9 & 74 & 75 & 75 & 74 & 74 & 74 \\
Qwen3-8B & 66 [60, 73] & 66 [53, 82] & 99.9 & 67 & 68 & 68 & 67 & 67 & 67 \\
Qwen3-14B & 72 [66, 78] & 72 [51, 94] & 99.8 & 74 & 76 & 75 & 74 & 74 & 74 \\
Granite-8B & 65 [57, 73] & 65 [35, 79] & 100.0 & 75 & 75 & 75 & 74 & 77 & 71 \\
\bottomrule
\end{tabular}
\end{table*}

\paragraph{Uncertainty from the products}
Each market has only three products. Resampling products as well as families widens the intervals considerably (Table~\ref{tab:robust}; at 768 tokens, 10--59\% for Phi-mini and 67--99\% for DeepSeek-8B), but every upper bound stays below the calculator, and arithmetic stays at most 15\% of the gap for five models. In a landscape bootstrap that resamples both construction and held-out auctions, the calculator beats every model in at least 96\% of draws (99.8\% or more for all but DeepSeek-8B), DeepSeek-8B stays best and Phi-mini worst in every draw.

\paragraph{Scoring curve}
Under the first-price equilibrium curve, gap recovered at 768 tokens is 40--85\%, against 42--85\% under the empirical curve; Granite-8B moves most between the two curves, from 65\% to 74\%. Under the first-price curve it moves by at most 4 points when the value distribution is gamma or Weibull instead of lognormal, when the bid increment is coarser, or when the number of bidders is one lower or higher (Table~\ref{tab:robust}).

\paragraph{Parser tolerance}
With the matcher's tolerance at 1\% or 5\% instead of 2\%, no share in Table~\ref{tab:splitebay} moves by more than 11 points.

\paragraph{Sampling and hardware}
Five sampled seeds move $q$ by at most 0.017, and the highest-bid lean holds under sampling (Section~\ref{sec:wrong}). Rerunning on different GPUs and with vLLM instead of HF transformers reproduces 94--100\% of saved letters and moves $q$ by at most 0.027.

\section{Discussion}

\paragraph{What operators can do}
Three practical lessons follow. First, do not cap reasoning tightly on decisions of this size: the tokens that finish the list cost a fraction of a cent, far below the profit at stake. Second, read the answer only after the model has closed its reasoning block; asking for it while the block is open can silently replace the trace's conclusion. Third, keep the reasoning trace and check it: a reader that only extracts the per-option numbers the model wrote, and flags disagreement with the answer, catches most of the costly errors, needs no calculator to raise the flag, and transferred across markets without retuning.

\paragraph{What belongs to the models}
Two behaviors survive the protocol changes we tried. Models pass over the option their own trace ranked best, with the block closed and without forcing, and their wrong answers lean toward the largest bid, with the block closed and under sampling. The lean resembles a pull toward the option most likely to win, at the expense of margin. In a pay-your-bid market that pull is costly to the bidder and profitable to the seller: against rivals in first-price equilibrium, moving the models' answers to the calculator's bids would cut seller revenue by \$13k--\$55k per 1{,}000 decisions. This matters for anyone deploying such agents on a bidder's behalf, and it gives a platform that earns from bids an incentive the bidder does not share.

\paragraph{Why we expect a lower bound}
Our task removes everything that makes bidding hard: the estimates are given, the menu is short and the objective is stated. Real agents also weigh budgets, pacing, competition and goals the calculator cannot represent. We expect losses in those settings to be at least as large, although we have not measured them.

\section{Limitations}
\begin{itemize}
\item \textbf{Markets.} Win curves come from eBay closing prices and one simulated advertising market; real ad-exchange logs are untested, and each market has only three products.
\item \textbf{Auction rules.} We score pay-your-bid profit on both markets and do not model eBay's proxy rule or AuctionNet's generalized second-price payments and slot exposure.
\item \textbf{Models.} We test open models of 4B to 14B parameters. Whether larger or frontier models also pass over and lean toward the highest bid is untested. Granite-8B was not run with a 4{,}096-token cap.
\item \textbf{Decoding.} The main protocol is greedy. Sampling moves $q$ little and keeps the highest-bid lean, but we have only the sampled letters, not the sampled traces, so the gap split under sampling is untested.
\item \textbf{Tools.} The calculator is an oracle; we do not test a model that calls a tool and can fail to.
\item \textbf{Parsing.} The split relies on code, not human labels. Lower and upper bounds on arithmetic bracket the uncertainty, and the two methods agree in 86--95\% of options where both find a value.
\item \textbf{Review cost.} The audit's override uses the calculator's bid, and we do not price the human or automated review a real operator would perform.
\end{itemize}

\section{Conclusion}
For small open reasoning models choosing a bid from a landscape, the loss is mostly not arithmetic. It comes from the token cap, from the readout prompt, which closing the reasoning block largely fixes, and from the final choice. Two findings belong to the models rather than to our protocol: passing over the best option the trace wrote, and a lean toward the highest bid that is about size, not position. Flagging answers that disagree with the trace's own numbers, and overriding them with the calculator's bid, recovers most of the loss, and the flag itself needs no calculator.

\section*{Data and ethics}
The eBay data are a public research dataset of completed listings. Our scenarios use only prices, not the public bidder handles in the source, and we do not redistribute the raw file. AuctionNet is simulated and distributed under the Apache 2.0 license; we release only derived scenario files with attribution. No human participants were involved.

\bibliographystyle{ACM-Reference-Format}
\bibliography{references}

\appendix
\FloatBarrier

\section{Prompts}
\label{app:prompts}

\paragraph{Bidding prompt} The options are listed in the randomized order of the decision's seed, labelled A, B, \ldots
\begin{lstlisting}
You are a bidder in a first-price auction.
Your private value is $<v>.

Available bids are shown in randomized order:
Option A: bid $<b_1>; estimated win probability <p_1>
Option B: bid $<b_2>; estimated win probability <p_2>
...

For any bid b, expected profit under the supplied estimate is
(private value - b) * estimated win probability.
Choose one listed bid. Do not infer the answer from option position.
End exactly with: FINAL BID: <number>
\end{lstlisting}

\paragraph{Readout (paper protocol)} Appended directly after the reasoning tokens, with the reasoning block left open; the most likely letter among the menu's letters is taken.
\begin{lstlisting}
Use the reasoning above and select the best displayed option. Return one option letter only. FINAL OPTION:
\end{lstlisting}

\paragraph{Readout (closed block)} The model's end-of-thinking marker, then the answer cue.
\begin{lstlisting}
<think> models:  "\n</think>\n\nFinal answer: Option"
Granite-8B:      "\n\nHere is my response:\nFinal answer: Option"
(if the marker already appears in the trace: "\n\nFinal answer: Option")
\end{lstlisting}

\paragraph{Prompting variant} Inserted before the last line of the bidding prompt.
\begin{lstlisting}
End by listing the expected profit of each option, then answer with the option that has the largest.
\end{lstlisting}

\FloatBarrier
\section{Additional results}
\label{app:extra}

\subsection{Full regression}
Table~\ref{tab:regfull} lists every coefficient of the passing-over regression on both markets, and Table~\ref{tab:regmodel} the per-model slopes from the eBay regression with model interactions. The highest-bid effect has the same sign for every model and is strongest for Qwen3-8B and Qwen3-14B. The position effect differs in direction across models: Phi-mini passes over a best option listed late far more often, while Granite-8B does the opposite. A finished list and a larger relative profit gap both make passing over less likely on both markets, and a 7-option menu makes it more likely.

\begin{table}[h!]
\caption{All coefficients of the passing-over regression (odds ratios [95\% interval], errors clustered by scenario; reference model Qwen3-8B). On eBay the reference product is Cartier watches; on AuctionNet it is high-intent impressions.}
\label{tab:regfull}
\footnotesize
\centering

\begin{tabular}{lcc}
\toprule
Term & eBay & AuctionNet \\
\midrule
Intercept & 0.68 [0.47, 0.99] & 0.49 [0.34, 0.72] \\
Relative profit gap & 0.36 [0.13, 0.99] & 0.09 [0.03, 0.26] \\
Position of best option & 1.93 [1.42, 2.61] & 2.48 [1.84, 3.35] \\
$K=7$ & 1.89 [1.48, 2.41] & 1.95 [1.57, 2.43] \\
log value / product median & 1.08 [0.77, 1.51] & 3.10 [1.98, 4.83] \\
Best option is the highest bid & 0.36 [0.28, 0.46] & 0.29 [0.22, 0.38] \\
List finished & 0.38 [0.29, 0.51] & 0.43 [0.32, 0.58] \\
Model: DeepSeek-8B & 0.49 [0.35, 0.68] & 0.31 [0.22, 0.43] \\
Model: Phi-mini & 5.92 [4.27, 8.21] & 5.92 [4.36, 8.04] \\
Model: Qwen3-4B & 1.03 [0.80, 1.32] & 0.71 [0.54, 0.92] \\
Model: Qwen3-14B & 0.74 [0.58, 0.94] & 0.70 [0.54, 0.91] \\
Model: Granite-8B & 3.04 [2.12, 4.34] & 3.07 [2.29, 4.11] \\
eBay product: Palm Pilot & 0.50 [0.37, 0.68] & -- \\
eBay product: Xbox & 0.48 [0.35, 0.65] & -- \\
AuctionNet product: low-intent & -- & 0.94 [0.73, 1.22] \\
AuctionNet product: mid-intent & -- & 1.03 [0.81, 1.32] \\
\bottomrule
\end{tabular}
\end{table}

\begin{table}[h!]
\caption{Per-model slopes from the eBay regression with model interactions, odds ratios [95\% interval].}
\label{tab:regmodel}
\small
\centering

\begin{tabular}{lcc}
\toprule
Model & Best is highest & Position \\
\midrule
DeepSeek-8B & 0.29 [0.10, 0.80] & 1.33 [0.63, 2.78] \\
Phi-mini & 0.55 [0.34, 0.89] & 17.79 [9.27, 34.13] \\
Qwen3-4B & 0.31 [0.16, 0.61] & 1.69 [0.88, 3.26] \\
Qwen3-8B & 0.03 [0.00, 0.19] & 4.40 [2.06, 9.41] \\
Qwen3-14B & 0.04 [0.01, 0.29] & 2.22 [1.04, 4.75] \\
Granite-8B & 0.51 [0.31, 0.83] & 0.15 [0.08, 0.27] \\
\bottomrule
\end{tabular}
\end{table}

\subsection{Token use without forcing}
Table~\ref{tab:unforced} gives token use for the unforced runs. Under a 1{,}024-token cap, 70--87\% of traces reach the cap for DeepSeek-8B, Phi-mini, Qwen3-4B and Qwen3-8B, and 46\% for Qwen3-14B, which is why the 1{,}024 run sits between the forced 768-token run and the 4{,}096 run. Under a 4{,}096-token cap the models stop on their own almost always; Phi-mini reaches the cap in 6.8\% of decisions. Granite-8B stops early, at about 560 tokens on average, and never reaches the 1{,}024 cap.

\begin{table}[h!]
\caption{Token use without forcing: mean or median reasoning tokens and the share of decisions that reach the cap (\%).}
\label{tab:unforced}
\footnotesize
\centering

\begin{tabular}{lcccccccc}
\toprule
& \multicolumn{4}{c}{eBay} & \multicolumn{4}{c}{AuctionNet} \\ \cmidrule(lr){2-5}\cmidrule(lr){6-9} & \multicolumn{2}{c}{1,024 cap} & \multicolumn{2}{c}{4,096 cap} & \multicolumn{2}{c}{1,024 cap} & \multicolumn{2}{c}{4,096 cap} \\ Model & Mean tok. & At cap & Median tok. & At cap & Mean tok. & At cap & Median tok. & At cap \\
\midrule
DeepSeek-8B & 1008 & 86 & 1327 & 0.6 & 1014 & 87 & 1324 & 0.9 \\
Phi-mini & 1002 & 87 & 1364 & 6.8 & 1022 & 97 & -- & -- \\
Qwen3-4B & 950 & 70 & 1390 & 1.1 & 970 & 74 & 1474 & 0.0 \\
Qwen3-8B & 969 & 73 & 1226 & 0.6 & 999 & 79 & 1265 & 0.2 \\
Qwen3-14B & 870 & 46 & 980 & 0.0 & 929 & 52 & 1047 & 0.0 \\
Granite-8B & 560 & 0 & -- & -- & 562 & 0 & -- & -- \\
\bottomrule
\end{tabular}
\end{table}

\subsection{Stated answers}
Table~\ref{tab:stated} reports the stated-answer parser. The Qwen models state an explicit choice in only 27--47\% of 768-token traces, because many are cut off before they get there, but where they do, the choice agrees with the letter in 96--98\% of cases. Phi-mini states a choice in 60\% of traces, and that choice is right 94\% of the time while its letter is right 55\% of the time. Granite-8B's stated choice and its letter agree, and both are often wrong, consistent with a decision error rather than a readout error.

\begin{table}[h!]
\caption{The stated-answer parser on eBay (\%). Coverage: traces in which the parser finds a stated choice. Agrees: that choice equals the letter. Stated right and Letter right: share of parsed traces in which the stated choice, or the letter, is the calculator's bid.}
\label{tab:stated}
\footnotesize
\centering

\begin{tabular}{lcccccc}
\toprule
& \multicolumn{4}{c}{768 forced tokens} & \multicolumn{2}{c}{1,024 run} \\ \cmidrule(lr){2-5}\cmidrule(lr){6-7} Model & Coverage & Agrees & Stated right & Letter right & Coverage & Agrees \\
\midrule
DeepSeek-8B & 82 & 89 & 85 & 94 & 96 & 94 \\
Phi-mini & 60 & 57 & 94 & 55 & 92 & 54 \\
Qwen3-4B & 37 & 96 & 98 & 97 & 67 & 99 \\
Qwen3-8B & 27 & 98 & 98 & 99 & 62 & 99 \\
Qwen3-14B & 47 & 98 & 98 & 98 & 74 & 98 \\
Granite-8B & 74 & 92 & 57 & 53 & 73 & 99 \\
\bottomrule
\end{tabular}
\end{table}

\subsection{Agreement between the two trace readers}
Table~\ref{tab:codecheck} checks the code that reads the traces at 768 forced tokens. Where the matcher and the text reader both find a value for an option, the reader's number equals the true expected profit in 86--95\% of options. At most 13\% of traces contain a miscomputed value for five models, against 25\% for Granite-8B, in line with Granite-8B's larger arithmetic share.

\begin{table}[h!]
\caption{Agreement between the two pieces of trace-reading code at 768 forced tokens on eBay. Options written: options for which the trace states a value. Found by both: the matcher and the reader both find it. Reader = truth: the reader's number equals the true expected profit (\%). Last column: traces in which at least one written value is miscomputed (\%).}
\label{tab:codecheck}
\footnotesize
\centering

\begin{tabular}{lcccc}
\toprule
Model & Options written & Found by both & Reader = truth & Traces with a miscomputed value \\
\midrule
DeepSeek-8B & 3,752 & 3,734 & 95 & 3 \\
Phi-mini & 3,529 & 3,470 & 92 & 8 \\
Qwen3-4B & 3,221 & 3,152 & 93 & 10 \\
Qwen3-8B & 3,028 & 2,943 & 93 & 13 \\
Qwen3-14B & 3,228 & 3,137 & 92 & 13 \\
Granite-8B & 3,511 & 2,996 & 86 & 25 \\
\bottomrule
\end{tabular}
\end{table}

\subsection{AuctionNet data and products}
Table~\ref{tab:andata} describes the three AuctionNet products and Table~\ref{tab:ancat} the gap recovered at 768 tokens by product. The ranking of models is broadly stable across products, except that Phi-mini recovers nothing on mid-intent impressions with the paper's readout.

\begin{table}[h!]
\caption{AuctionNet products. Opportunities are ad impressions from periods 7--9 (construction) and 10 (held-out); values and bids are per 1{,}000 impressions. Grid points: candidate bids between the 2nd and 98th percentile of competing prices.}
\label{tab:andata}
\footnotesize
\centering

\begin{tabular}{lccccc}
\toprule
Product & Construction opp. & Held-out opp. & Values & Bid step & Grid points \\
\midrule
High-intent & 499,978 & 183,975 & 102--206 & 2 & 56 \\
Mid-intent & 499,977 & 132,479 & 92--198 & 2 & 60 \\
Low-intent & 499,978 & 183,525 & 74--176 & 2 & 66 \\
\bottomrule
\end{tabular}
\end{table}

\begin{table}[h!]
\caption{AuctionNet gap recovered at 768 forced tokens by product (\%, point estimates).}
\label{tab:ancat}
\footnotesize
\centering

\begin{tabular}{lccc}
\toprule
Model & High-intent & Mid-intent & Low-intent \\
\midrule
DeepSeek-8B & 91 & 95 & 90 \\
Phi-mini & 33 & 0 & 25 \\
Qwen3-4B & 83 & 74 & 82 \\
Qwen3-8B & 75 & 66 & 62 \\
Qwen3-14B & 83 & 76 & 87 \\
Granite-8B & 62 & 85 & 74 \\
\bottomrule
\end{tabular}
\end{table}

\subsection{Stakes per decision}
Table~\ref{tab:ustar} gives the distribution of the best expected profit across eBay scenarios. Even the smallest stake (\$1.75) is far above the cost of the reasoning tokens (Table~\ref{tab:cost}).

\begin{table}[h!]
\caption{Best expected profit per eBay scenario, $\max_b U(b)$, by percentile. It sets the stakes against which the cost of reasoning tokens is compared (Table~\ref{tab:cost}).}
\label{tab:ustar}
\footnotesize
\centering

\begin{tabular}{ccccccc}
\toprule
Min & 5th & 25th & Median & 75th & 90th & Max \\
\midrule
\$1.75 & \$4.13 & \$15.49 & \$48.33 & \$136.00 & \$840.24 & \$1,521.22 \\
\bottomrule
\end{tabular}
\end{table}

\end{document}